\documentclass{article}
\usepackage{spconf,amsmath,graphicx}
\usepackage{multirow}
\usepackage{color}
\usepackage{hyperref}
\hypersetup{colorlinks=true}
\usepackage{booktabs}
\usepackage[final]{changes}

\usepackage{bibspacing}
\usepackage{color}
\definecolor{darkspringgreen}{rgb}{0.39, 0.65, 0.27}

\title{Scaling Up Deliberation for Multilingual ASR}
\name{BLIND}
\address{BLIND}
\name{Ke Hu, Bo Li, Tara N. Sainath}
\address{Google LLC, USA \\
\fontsize{9}{9}\selectfont\ttfamily\upshape
\{huk,boboli,tsainath\}@google.com}
\begin{document}
\ninept
\maketitle
\begin{abstract}
Multilingual end-to-end automatic speech recognition models are attractive due to its simplicity in training and deployment. Recent work on large-scale training of such models has shown promising results compared to monolingual models. However, the work often focuses on multilingual models themselves in a single-pass setup. In this work, we investigate second-pass deliberation for multilingual speech recognition. Our proposed deliberation is multilingual, i.e., the text encoder encodes hypothesis text from multiple languages, and the decoder attends to multilingual text and audio. We investigate scaling the deliberation text encoder and decoder, and compare scaling the deliberation decoder and the first-pass cascaded encoder. We show that deliberation improves the average WER on 9 languages by 4\% relative compared to the single-pass model. By increasing the size of the deliberation up to 1B parameters, the average WER improvement increases to 9\%, with up to 14\% for certain languages. Our deliberation rescorer is based on transformer layers and can be parallelized during rescoring.
\end{abstract}
\section{Introduction}
\label{sec:intro}

There has been growing interest in developing multilingual end-to-end (E2E) automatic speech recognition (ASR) models in the past few years \cite{watanabe2017language,kim2018towards,zhou2018multilingual,kannan2019large,hou2020large,zhu2020multilingual,zhou2022configurable}. Previous work on multilingual E2E models has explored different model structures such as connectionist temporal classification (CTC) models \cite{kim2018towards}, attention-decoder based models \cite{watanabe2017language, zhou2018multilingual,hou2020large}, and streaming models \cite{kannan2019large, zhu2020multilingual}. While the previous research mainly focuses on various E2E model structures, recent large-scale data sets have motivated work in developing giant multilingual ASR models \cite{pratap2020massively, adams2019massively, li2021scaling, li2022language, li2022massively}. For example, \cite{li2022language} has proposed a truly multilingual on-device transducer model for 9 languages without explicitly using language information. By increasing the model capacity to 1B parameters, \cite{li2022language} shows that the model performs generally better than monolingual models for all languages. In another large-scale training, \cite{pratap2020massively} builds a multilingual E2E model up to 1B size, and the authors have improved quality of all variants of the multilingual model \cite{pratap2020massively}. However, the training data used in \cite{pratap2020massively} is relatively small compared to \cite{li2022language}. Although the aforementioned studies pool all languages together and train from scratch, \cite{li2022massively} has proposed a life-long learning strategy for large-scale multilingual model training. By adding languages incrementally and increasing the model size up to 1B, \cite{li2022massively} shows that a lifelong learning strategy is more effective than training from scratch.

Recently, two-pass ASR models have been shown to further improve first-pass model performance \cite{sainath2019twopass, hu2020deliberation,variani2020neural,wang2022deliberation}. While increasing the capacity of the multilingual models themselves improves recognition quality, it is unclear whether a second-pass model works in a multilingual setup and how much it can benefit from a bigger size. In \cite{gaur2022multilingual}, the authors have investigated both multilingual and monolingual second-pass rescoring for a 6-language multilingual model and obtained significant improvements in both scenarios. However, the biggest model used is around 300M and it is unclear how the second-pass performance changes as capacity increases. Traditional shallow fusion \cite{kannan2018analysis} or neural language model (LM) rescoring \cite{sainath2021efficient} are mostly language dependent, i.e. each language has a monolingual LM for post-processing.

Deliberation networks are one type of two-pass models, and have achieved state-of-the-art results in monolingual ASR for English \cite{xia2017deliberation, hu2020deliberation, mavandadi2021deliberation, wang2022deliberation, hu2022improving}. Deliberation is a two-pass ASR model where the first-pass is usually a transducer \cite{hu2020deliberation} generating first-pass text hypotheses, and in the second pass, a two-source attention is used to attend to both first-pass hypotheses and audio encoder outputs for redecoding or rescoring \cite{hu2020deliberation}. Note that in the monolingual scenario, the first-pass hypotheses are in a single language to deliberate \cite{hu2020deliberation}. It is unknown how deliberation performs in a multilingual scenario where hypothesis texts are in different languages and audio inputs are multilingual as well. 

In this work, we investigate deliberation for multilingual second-pass rescoring. We extend the model from \cite{li2022language} and add a cascaded encoder \cite{narayanan2021cascaded} as the first-pass multilingual model. A multilingual deliberation decoder is then used for second-pass rescoring. The deliberation decoder consists of a multilingual text encoder and a transformer-based multi-source attention decoder \cite{hu2021transformer}. We show that, in a truly multilingual setup without explicitly using any language information, deliberation improves the multilingual first-pass by 4\% in terms of average WER, despite both hypothesis text and audio encoder outputs are from multiple languages. Second, we show that by scaling up deliberation text encoder and the deliberation decoder, further improvements are achieved. Specifically, it is more effective to increase the width and depth of the deliberation decoder than the text encoder or the first-pass cascaded encoder. By increasing the multilingual deliberation up to 1B parameters, we improve the average WER by 9\% relative for 9 languages. The improvement is uniform for all languages and up to 14\% for certain languages. Compared to a first-pass cascaded encoder in a similar size (1B), the deliberation model has a relative average WER improvement of 4\% and per-language improvement up to 8\%.  As far as we know, this is the first work on large-scale training of deliberation for multilingual ASR.

\section{System Description}
\label{sec:system}

\subsection{Multilingual Deliberation}
\label{sec:scale_up}

Previous work shows that deliberation achieves the state-of-the-art results for monolingual models \cite{hu2020deliberation, hu2021transformer, hu2022improving}. However, it remains unknown how deliberation works for multilingual models. There are multiple choices in designing a deliberation model in a multilingual setup, e.g., one can use a per-language deliberation decoder after a first-pass model, or a single deliberation decoder for all languages. The former is more similar to a monolingual deliberation and may need explicit language information in inference, while the latter is truly multilingual and simpler to train and inference. In this work, we investigate a truly multilingual deliberation model, i.e. a single deliberation decoder for deliberation of multiple languages.

There are a few distinctions between multilingual deliberation and its monolingual versions: 1) The hypotheses from the first-pass model are in multiple languages, and it is unclear whether a single deliberation text encoder can model multilingual inputs, 2) A single deliberation decoder attends to both text and audio encodings in different languages, and 3) Does the deliberation also need increased capacity in the multilingual scenario (similar to \cite{li2021scaling}) to achieve comparable improvements in monolingual scenarios?

\begin{figure}[t]
  \centering
   \includegraphics[scale=0.4]{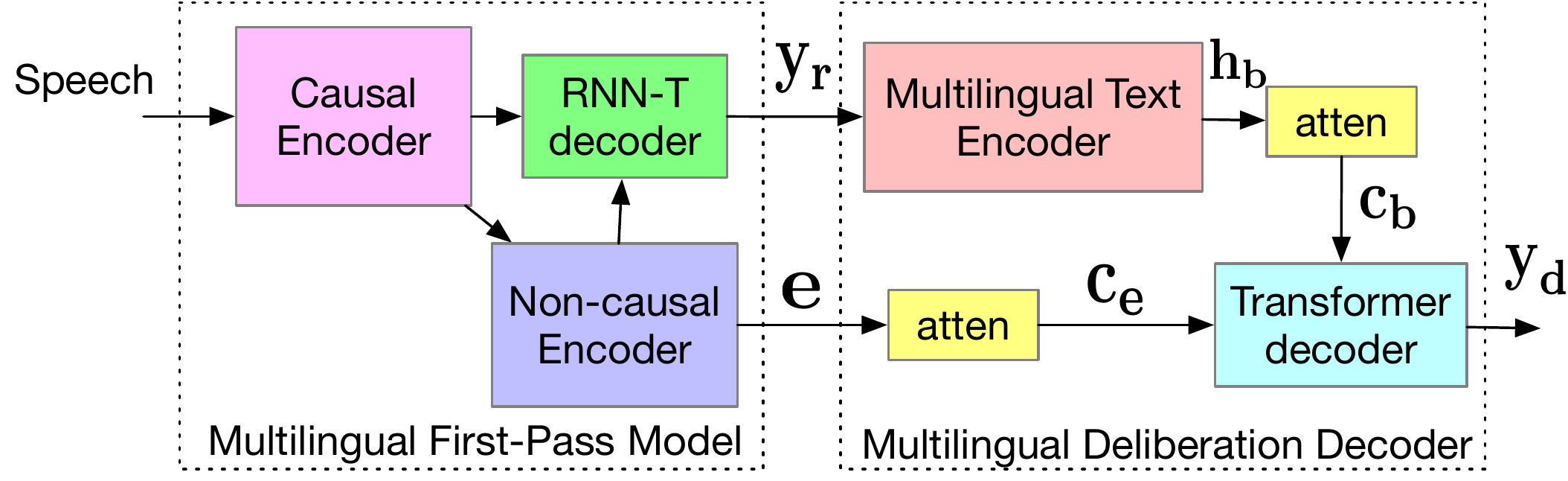}
   \caption{Multilingual deliberation based on a truly multilingual cascaded encoder model as the first pass.}
   \label{fig:multilingual_delib}
\end{figure}

We show the diagram of the proposed multilingual deliberation model in Fig. \ref{fig:multilingual_delib}. To achieve a truly multilingual system, we use a language agnostic multilingual model proposed in \cite{li2022language} as the first pass. The first-pass model does not require any any explicit language information in either encoder or decoder. Note that \cite{li2022language} only uses left context for audio encoders, and we use a non-causal cascaded encoder \cite{narayanan2021cascaded} to leverage the audio right context. Note that our cascaded encoder is also multilingual and does not explicitly use any language information. For all languages, we sample the first-pass decoder softmax \cite{xia2017deliberation} to generate multilingual hypothesis outputs. To achieve a truly multilingual ASR system, our multilingual text encoder does not use any explicit language information and encodes text inputs of multiple languages using shared parameters. Usually a bidirectional LSTM \cite{hu2020deliberation} or a conformer encoder~\cite{gulati2020conformer} is used as the text encoder. Similarly, to achieve a truly multilingual rescorer, we use a single deliberation decoder based on transformer layers \cite{hu2021transformer}. No explicit language information is used during rescoring. The multilingual deliberation decoder attends to both the cascaded encoder (non-causal) output ($\bf{e}$) and hypotheses ($\bf{y_r}$) from the non-causal encoder. 

In an ASR task of 9 languages, we use 16,384 wordpieces as the output vocabulary for both first-pass multilingual model and the second-pass deliberation. The wordpicece model is generated on the text transcriptions of our training data by pooling all the languages together. We use a word count threshold of 20 to drop uncommon words when building the wordpiece model. For Chinese, due to the ambiguity in word segmentation, we simply use characters instead words. While for Japanese, we use MeCab \cite{kudo2005mecab} to segment the text into words. 
During training, we similarly pool all the data together and sample each batch according to the natural distribution. We freeze the first-pass model and only the deliberation network is optimized using a cross entropy loss between the output of the network and the ground truth transcripts.

\subsection{Scaling Up Deliberation}
\cite{li2021scaling} shows that scaling up the number of model parameters is an efficient way to increase modeling capacity for multilingual ASR. Inspired by \cite{li2021scaling}, we scale up the deliberation architecture in different ways to increase the capacity for multilingual rescoring. As shown in Fig. \ref{fig:multilingual_delib}, there are two major components in the deliberation decoder: 1) Multilingual text encoder, and 2) Multilingual two-source transformer decoder. In this work, we empirically study the effect of increasing the capacity of the two components. We experiment with increasing both the width or the depth of the text encoder for modeling the hypotheses in multiple languages. While \cite{li2022language} finds that increasing encoder in a transducer model is effective in increasing model capacity, we increase the attention-based decoder up to 1B parameters and compare that to \cite{li2022language} with a large cascaded encoder. We also research alternatives such as scaling up the non-causal cascaded encoder first and then add deliberation or scaling up the deliberation decoder alone. Training large-scale model is challenging and we use Adafactor \cite{shazeer2018adafactor} and repeated layers \cite{shen2019lingvo} for reduce high-bandwidth memory use. Our models are trained in Tensorflow~\cite{abadi2016tensorflow} using the Lingvo framework~\cite{shen2019lingvo} on  4$\times$4$\times$8 Tensor Processing Units (TPU) v4 slices \cite{tpuv42022} with a global batch size of 4,096, and optimized using synchronized stochastic gradient descent. We cap the gradient norm change of a parameter to 5.0 to stabilize training.

\section{Experimental Details}

\subsection{Data}
Our multilingual training data consists of Voice Search speech from 9 languages in different locales, including English (USA), Chinese, French, German, Japanese, Spanish (USA), Spanish (Spain), Italian and English (UK) (see Table \ref{tab:data} for more details). They consist of around 142M utterances and the total duration is around 214K hours. The training data is anonymized and human transcribed. The per-language training data ranges from 6M to 34M. For each language, we use a test set with utterances sampled from the Voice Search traffic for each language and they range from 23.8K to 300K utterances. The test set does not overlap with the training set, and they are also anonymized and human transcribed.

\begin{table}[t]
\centering
\vspace{-0.1in}
\begin{tabular}{llrr}
\toprule
Locale & Language & Counts (M) & Hours (K)\\
\midrule
\midrule
en-US & English (USA) & 34.0 & 52.6 \\
zh-TW & Chinese & 16.6 & 22.0 \\
fr-FR & French & 16.5 & 23.7 \\
de-DE & German &  15.3 & 23.4 \\
ja-JP & Japanese & 14.9 & 20.2 \\
es-US & Spanish (USA) & 14.2 & 23.8 \\
es-ES & Spanish (Spain) & 12.9 & 20.1 \\
it-IT & Italian & 11.8 & 19.8 \\
en-GB & English (UK) & 6.0 & 8.6 \\
\midrule 
\midrule
Total & & 142.2 & 214.2 \\
\end{tabular}
\caption{Training data for 9 languages. Utterance counts are in millions (M) and duration is in thousand (K) of hours.}
\label{tab:data}
\end{table}

\subsection{Model Description}

We use a baseline multilingual model similar to \cite{li2022language} which is language agnostic. The baseline model
consists of a 12-layer causal conformer encoder and a 5-layer non-causal cascaded encoder. The causal encoder includes two blocks separated by a stacking layer. The first block consists of a input projection layer and 3 conformer layers.
The stacking layer concatenates two neighboring encodings in time to form a 60-ms frame rate. Then the second block starts with a 1024-dim conformer layer, and then a projection layer to reduce the model dimension back to 512 for the rest of the layers. Note that the causal conformer layers uses causal convolution and left-context attention
and is thus strictly causal. Secondly, the non-causal layers are cascaded \cite{narayanan2021cascaded} to the causal encoder output. The 5 layers of non-causal conformer layers have the same dimension of 512 as the causal encoder, and a total right-context of 0.9 sec. The outputs from causal and non-causal encoders are sampled with probabilities of 0.4, and 0.6, respectively, and fed to a transducer decoder during training \cite{li2022language}.

The transducer decoder consists of a prediction network and a joint network \cite{graves2012sequence}. We use a 2-layer 2048D LSTM as the prediction network whose output is projected to 640 dimension for efficiency. The joint network is a single feed-forward layer of 640 units. A softmax is finally used to predict 16,384 wordpieces. We generate the wordpieces using mixed transcripts pooled from all languages. In summary, the baseline multilingual transducer model has a total of 173M parameters.

Our transformer-based deliberation decoder attends to non-causal encoder outputs and hypotheses from the first-pass multilingual model decoded using the non-causal encoder. For efficiency, we sample from the softmax outputs to generate a single hypothesis token for each frame in both training and inference. In the deliberation decoder, we use a 2-layer 2048D bidirectional LSTM as the text encoder, totaling around 30M parameters. The transformer decoder consists of 4 transformer decoder layers (similar to~\cite{hu2021transformer}), where each layer has 2,048 hidden units followed by 512-dimensional projection. A 16,384-dimensional softmax is then used to predict the same wordpieces as the first-pass multilingual model. The deliberation decoder has around 62M parameters. In inference, we use beam search to decode the first-pass model and generate 8 hypotheses, and use the deliberation decoder to rescore them. Note that the forward pass from the RNN-T decoder to the multilingual text encoder is still done by sampling. 

For the input speech, it is divided by 32-ms windows with a frame rate of 10 ms. 80D log-Mel filterbank features are extracted from each frame and then stacked together from 3 continuous frames to form a 240D input vector. These input vectors are further downsampled to have a 30-ms frame rate. We use SpecAug \cite{park2019specaugment} to improve model robustness against noise. Two frequency masks with a maximum length of 27 and two time masks with a maximum length of 50 are used.

\subsection{Training}
As mentioned in Sect. \ref{sec:scale_up}, our models are trained in Tensorflow~\cite{abadi2016tensorflow} using the Lingvo framework~\cite{shen2019lingvo} on 4$\times$4$\times$8 Tensor Processing Units (TPU) v4 slices. For multilingual deliberation models, we use a linear learning rate schedule which will warm up in the first 32K steps and then stay constant. We use a variant of Adam optimizer with parameters $\beta_1=0.9$ and $\beta_2=0.999$ and cap the gradient norm change of a parameter to 5.0 to stabilize training. For training large multilingual cascaded encoder models, we use a transformer learning rate schedule with peak learning rate of 1.8e-3 and a 32K-step warm-up. For the 1B multilingual cascaded encoder model, we use Adafactor \cite{shazeer2018adafactor} for memory efficiency. Exponential moving average~\cite{polyak1992acceleration} has been used by all models to mitigate model parameter fluctuations.

\section{Results}
\label{sec:results}

\subsection{Multilingual Deliberation}

\subsubsection{Deliberation using Causal Encoder}
Deliberation has been shown to significantly improve monolingual English first-pass models \cite{hu2020deliberation, hu2021transformer, hu2022transducer} but its effectiveness remains unknown to multilingual models. When one applies deliberation to a transducer model with cascaded encoder \cite{li2022language}, the deliberation decoder can attend to either hypotheses decoded using the causal encoder, or the non-causal encoder. In Table \ref{tab:delib_causal}, we deliberate on the causal decoding results for better latency. We show in Table \ref{tab:delib_causal} that deliberation (\texttt{E0}) improves the baseline (\texttt{B0}) average WER by around 5\% relative. We note the improvement is uniform and significant for all languages. Note that our deliberation decoder does not explicitly use any languages information and is thus truly multilingual. The improvement shows that deliberation works effectively for multilingual models, and the improvement in Table \ref{tab:delib_causal} presumably comes from both bidirectional encoding of the hypothesis and a second-pass rescoring using both hypotheses and acoustic encoding.

\begin{table}[t]
\centering
\begin{tabular}{lrr}
    \toprule
    & \multicolumn{1}{c}{B0} & \multicolumn{1}{c}{E0} \\ \cline{2-3}
    \multirow{2}{*}{Model} & \multirow{2.2}{*}{\shortstack{ML \\ Causal}} & \multirow{2.2}{*}{\shortstack{ML \\ Delib.}} \\
    & & \\ \cline{1-3}
    \multirow{1.2}{*}{Size} & \multirow{1.2}{*}{143M} & \multirow{1.2}{*}{208M} \\ \midrule \midrule
    en-US & 6.3 & \textbf{6.1} \\
    fr-FR & 12.9 & \textbf{12.0} \\
    es-US & 7.3 & \textbf{6.8} \\
    en-GB & 6.3 & \textbf{6.1} \\
    es-ES & 7.0 & \textbf{6.8} \\
    ja-JP & 12.1 & \textbf{11.2} \\
    zh-TW & 5.3 & \textbf{5.0} \\
    de-DE & 13.6 & \textbf{13.1} \\
    it-IT & 8.1 & \textbf{7.6} \\ \hline
    \multirow{1.2}{*}{Avg. WER} & \multirow{1.2}{*}{8.77} & \multirow{1.2}{*}{\textbf{8.30}} \\[0.3ex] \hline
\end{tabular}
\caption{WERs (\%) between a multilingual (ML) transducer using a causal encoder and deliberation.}
\label{tab:delib_causal}
\end{table}

\subsubsection{Deliberation using Cascaded Encoder}
 While a causal encoder is used for decoding in Table \ref{tab:delib_causal}, a cascaded encoder \cite{narayanan2021cascaded} further improves recognition by using right-contexts in the audio encoder. In Table \ref{tab:delib}, we use deliberation on the first-pass hypotheses decoded using a non-causal encoder (i.e., the cascaded encoder). We show that multilingual deliberation (\texttt{E1}) performs slightly better than the multilingual cascaded encoder baseline (\texttt{B1}). \texttt{E1} has a 4-layer 512D transformer decoder and a 2-layer bidirectional LSTM text encoder. The limited quality improvement may be due to the small size of the deliberation decoder as \cite{li2021scaling} shows that multilingual models perform better with large capacity. We thus increase the transformer decoder dimension to 640 (\texttt{E2}), and as shown in Table \ref{tab:delib}, the average WER improves correspondingly. Note that here we use a conformer text encoder for \texttt{E2} due to its efficiency \cite{hu2022transducer}. When we further increase the decoder depth to 8 and dimension to 720, the average WER improves to 7.39\%. This is around 4\% relative better than the multilingual cascaded encoder baseline (\texttt{E3} v.s. \texttt{B1}). Comparing \texttt{E3} and \texttt{B1}, we note that the cascaded encoder (\texttt{B1}) has a smaller size than \texttt{E3}. We will have a more systematic comparison in Sect. \ref{sec:comparison}.

\begin{table}[t]
\centering
\begin{tabular}{lcccc}
    \toprule
    & \multicolumn{1}{c}{B1} & \multicolumn{1}{c}{E1} & \multicolumn{1}{c}{E2} & \multicolumn{1}{c}{E3} \\ \cline{2-5}
    \multirow{2}{*}{Model} & \multirow{2.2}{*}{\shortstack{ML \\ Cascaded}} & \multirow{2.2}{*}{\shortstack{ML \\ Delib.}} & \multirow{2.2}{*}{\shortstack{640D \\ Trans. Dec.}} & \multirow{2.2}{*}{\shortstack{8L720D \\ Trans. Dec.}} \\
    & & & & \\ \cline{1-5}
    \multirow{1.2}{*}{Size} & \multirow{1.2}{*}{174M} & \multirow{1.2}{*}{239M} & \multirow{1.2}{*}{247M} & \multirow{1.2}{*}{300M} \\ \midrule \midrule 
    en-US & 5.5 & 5.4 & 5.3 & \textbf{5.1} \\
    fr-FR & 11.7 & 11.6 & 11.6 & \textbf{11.6} \\
    es-US & 6.6 & 6.6 & 6.5 & \textbf{6.5} \\
    en-GB & 5.5 & 5.5 & 5.4 & \textbf{5.2} \\
    es-ES & 6.4 & 6.2 & 6.2 & \textbf{6.1} \\
    ja-JP & 10.2 & 10.0 & 10.0 & \textbf{9.7} \\
    zh-TW & 4.6 & 4.6 & 4.5 & \textbf{4.4} \\
    de-DE & 12.2 & 12.1 & 12.1 & \textbf{11.8} \\
    it-IT & 6.6 & 6.3 & 6.3 & \textbf{6.1} \\ \hline
    \multirow{1.2}{*}{Avg. WER} & \multirow{1.2}{*}{7.70} & \multirow{1.2}{*}{7.59} & \multirow{1.2}{*}{7.54} & \multirow{1.2}{*}{\textbf{7.39}} \\[0.3ex] \hline
\end{tabular}
\caption{WERs (\%) between multilingual (ML) cascaded encoder and deliberation transformer decoders with different depth and width.}
\label{tab:delib}
\end{table}

\subsection{Scaling Up Text Encoder}

Next, we study the effect of increasing the text encoder of the deliberation model. Starting from \texttt{E1} in Table \ref{tab:delib}, we further increase the text encoder of the deliberation decoder to a 6L 1024D conformer (\texttt{E4} in Table \ref{tab:text_enc}). This improves the WER slightly compared to the baseline deliberation (\texttt{E1}). Note that \texttt{E1} has an bidirectional-LSTM text encoder and we have also tried using a conformer text encoder with a similar size (around 239M) and obtained similar results. However, as we further increase the size of the text encoder in Table \ref{tab:text_enc}, the WER starts to degrade. For example, when using a 11L 1024D (\texttt{E5}) or a 8L 2048D (\texttt{E6}) conformer text encoder, the model performs significantly worse. This actually confirms the results in \cite{hu2022improving} where the authors find that a large text encoder without pretraining introduces regression. Multilingual text encoder pretraining is not the scope of this paper, but in future we plan to research that to facilitate training of large text-encoder deliberation model. All the aforementioned text conformer encoders have four-token lookahead. 

\begin{table}[t]
\centering
\begin{tabular}{lccc}
    \toprule
    & \multicolumn{1}{c}{E4} & \multicolumn{1}{c}{E5} & \multicolumn{1}{c}{E6} \\ \cline{2-4}
    \multirow{2}{*}{Model} & \multirow{2.2}{*}{\shortstack{6L1024D \\Text Enc.}} & \multirow{2.2}{*}{\shortstack{11L1024D \\Text Enc.}} & \multirow{2.2}{*}{\shortstack{8L2048D \\Text Enc.}} \\
    & & & \\ \cline{1-4}
    \multirow{1.2}{*}{Size} & \multirow{1.2}{*}{300M} & \multirow{1.2}{*}{500M} & \multirow{1.2}{*}{1B} \\ \midrule \midrule 
    en-US & \textbf{5.3} & 5.8 & 5.7 \\
    fr-FR & \textbf{11.5} & 11.7 & 11.5 \\
    es-US & \textbf{6.4} & 6.7 & 6.6 \\
    en-GB & \textbf{5.4} & 5.7 & 5.6 \\
    es-ES & \textbf{6.2} & 6.4 & 6.4 \\
    ja-JP & \textbf{9.9} & 10.8 & 10.7 \\
    zh-TW & \textbf{4.5} & 5.0 & 5.1 \\
    de-DE & \textbf{12} & 12.6 & 12.7 \\
    it-IT & \textbf{6.3} & 6.6 & 6.7 \\ \hline
    \multirow{1.2}{*}{Avg. WER} & \multirow{1.2}{*}{\textbf{7.50}} & \multirow{1.2}{*}{7.92} & \multirow{1.2}{*}{7.89} \\[0.3ex] \hline
\end{tabular}
\caption{WERs (\%) by increasing the size of the multilingual deliberation text encoder.}
\label{tab:text_enc}
\end{table}

\begin{table}[h]
\centering
\begin{tabular}{lccc}
    \toprule
    & \multicolumn{1}{c}{E3} & \multicolumn{1}{c}{E7} & \multicolumn{1}{c}{E8} \\ \cline{2-4}
    \multirow{2}{*}{Model} & \multirow{2}{*}{\shortstack{ML \\ Delib.}} & \multirow{2}{*}{\shortstack{10L1080D \\Trans. Dec.}} & \multirow{2}{*}{\shortstack{32L1080D \\Trans. Dec.}} \\
    & & & \\ \cline{1-4}
    \multirow{1.2}{*}{Size} & \multirow{1.2}{*}{300M} & \multirow{1.2}{*}{500M} & \multirow{1.2}{*}{1B} \\ \midrule \midrule 
    en-US & 5.1 & 4.8 & \textbf{4.7} \\
    fr-FR & 11.6 & 11.4 & \textbf{11.3} \\
    es-US & 6.5 & 6.4 & \textbf{6.2} \\
    en-GB & 5.2 & 5.0 & \textbf{4.8} \\
    es-ES & 6.1 & 5.9 & \textbf{6.0} \\
    ja-JP & 9.7 & 9.0 & \textbf{8.8} \\
    zh-TW & 4.4 & 4.1 & \textbf{4.1} \\
    de-DE & 11.8 & 11.3 & \textbf{11.4} \\
    it-IT & 6.1 & 5.8 & \textbf{5.6} \\ \hline
    \multirow{1.2}{*}{Avg. WER} & \multirow{1.2}{*}{7.39} & \multirow{1.2}{*}{7.08} & \multirow{1.2}{*}{\textbf{6.99}} \\[0.3ex] \hline
\end{tabular}
\caption{WERs (\%) by increasing the size of the multilingual deliberation transformer decoder.}
\label{tab:delib_dec}
\end{table}

\subsection{Increasing Transformer Decoder Size}
Since increasing transformer decoder capacity seems to be effective in Table \ref{tab:delib}, we further increase the depth and width of the deliberation transformer decoder starting from \texttt{E3} in Table \ref{tab:delib}. As shown in Table \ref{tab:delib_dec}, we first increase the decoder to 10-layer 1080D transformer decoder (\texttt{E7}). Comparing to the 8L 720D decoder (\texttt{E3}), we improves the average WER by around 4\% relative: 7.39\% $\rightarrow$ 7.08\%. The improvement is uniform for all languages and as large as 7\% relative for ja-JP. If we further increase the depth of the decoder to 32 layers (\texttt{E8}, a total of 1B parameters), the improvement increases to around 5\%. This shows the effectiveness of increasing the decoder size in large-scale deliberation. We have also tried similar 1B model with larger dimension and shallower layers but did not manage to get improvement. Since we use a transformer decoder for rescoring here, the rescoring of different tokens can be done in parallel to improve latency.

\begin{table}[t]
\centering
\begin{tabular}{lcc}
    \toprule 
    & \multicolumn{1}{c}{E9} & \multicolumn{1}{c}{E7} \\ \cline{2-3}
    \multirow{2}{*}{Model} & \multirow{2}{*}{\shortstack{300M Cas. Enc.+\\ 4L512D Trans. Dec.}} & \multirow{2}{*}{\shortstack{B1+10L1080D \\Trans. Dec.}} \\
    & & \\ \cline{1-3}
    \multirow{1.2}{*}{Size} & \multirow{1.2}{*}{500M} & \multirow{1.2}{*}{500M} \\ \midrule \midrule 
    en-US & 5.1 & \textbf{4.8} \\
    fr-FR & 11.4 & \textbf{11.4} \\
    es-US & 6.4 & \textbf{6.4} \\
    en-GB & \textbf{4.9} & 5.0 \\
    es-ES & 5.9 & \textbf{5.9} \\
    ja-JP & 9.4 & \textbf{9} \\
    zh-TW & 4.3 & \textbf{4.1} \\
    de-DE & 11.6 & \textbf{11.3} \\
    it-IT & 5.8 & \textbf{5.8} \\ \hline
    \multirow{1.2}{*}{Avg. WER} & \multirow{1.2}{*}{7.20} & \multirow{1.2}{*}{\textbf{7.08}} \\[0.3ex] \hline
\end{tabular}
\caption{WER (\%) comparison between increasing cascaded encoder and increasing the transformer decoder.}
\label{tab:delib_or_cascade}
\end{table}

\subsection{Scaling Transformer Decoder V.S. Cascaded Encoder}
\label{sec:dec_vs_cascaded}

\begin{table*}[h]
\centering
\begin{tabular}{|ccccccccccccc|}
    \hline
    \multirow{2}{*}{ID} & \multirow{2}{*}{Model} & \multirow{2}{*}{Size} & \multicolumn{9}{c}{WER (\%)} & \multirow{2}{*}{\shortstack{Avg. \\WER (\%)}} \\ \cline{4-12}
     & & & \multirow{1.2}{*}{en-US} & \multirow{1.2}{*}{fr-FR} & \multirow{1.2}{*}{es-US} & \multirow{1.2}{*}{en-GB} & \multirow{1.2}{*}{es-ES} & \multirow{1.2}{*}{ja-JP} & \multirow{1.2}{*}{zh-TW} & \multirow{1.2}{*}{de-DE} & \multirow{1.2}{*}{it-IT} & \\[0.3ex] \hline\hline
    B1 & Multilingual Cascaded & 174M & 5.5 & 11.7 & 6.6 & 5.5 & 6.4 & 10.2 & 4.6 & 12.2 & 6.6 & 7.70 \\ \hline
    B2 & \hspace{1.5em}w/ 23L1152D Cas. Enc. & 1B & 4.9  & 11.7 & 6.7 & 5.2 & 6.1 & 9.1 & 4.2 & 11.6 & 5.8 & 7.26  \\ \hline
    E3 & \hspace{-1.5em}Multilingual Delib. & 300M & 5.1 & 11.6 & 6.5 & 5.2 & 6.1 & 9.7 & 4.4 & 11.8 & 6.1 & 7.39  \\ \hline
    E6 & \hspace{1em}w/ 8L2048D Text Enc. & 1B & 5.7 & 11.5 & 6.6 & 5.6 & 6.4 & 10.7 & 5.1 & 12.7 & 6.7 & 7.89 \\ \hline
    E8 & \hspace{2.4em}w/ 32L1080D Trans. Dec. & 1B & \textbf{4.7} & \textbf{11.3} & \textbf{6.2} & \textbf{4.8} & \textbf{6.0} & \textbf{8.8} & \textbf{4.1} & \textbf{11.4} & \textbf{5.6} & \textbf{6.99} \\ \hline
\end{tabular}
\caption{Comparison of baseline multilingual cascaded encoder and a 1B cascaded encoder with multilingual deliberation model with different sizes of text encoder or transformer decoder. Note that \texttt{B2}, \texttt{E6}, and \texttt{E8} have the same number of parameters (1B).}
\label{tab:wer}
\end{table*}

An alternative to increasing the deliberation decoder is to increase the size of the first-pass model. Previous results in \cite{li2022language} show that it is effective to increase the encoder of a transducer to 1B. Therefore, in Table \ref{tab:delib_or_cascade}, we first increase the cascaded encoder to 5L 1152D (total size 300M) and then add a 4L 512D transformer decoder (\texttt{E9}) to reach a total size 500M. Compared to the baseline cascaded encoder (\texttt{B1} in Table \ref{tab:delib}), \texttt{E9} improves the WER from 7.70\% to 7.20\%. However, if we add all the model capacity only to the deliberation decoder (\texttt{E7} with a 10L 1080D transformer decoder), the WER is improved to  7.08\%. \added{This shows that increasing the size of the deliberation decoder may be more effective than the size of the cascaded encoder.} We will also compare 1B deliberation decoder with a 1B cascaded encoder in Sect. \ref{sec:comparison}. We note that an exhaustive search of an optimal way of assigning model capacity to the cascaded encoder and deliberation transformer decoder is out of the scope of this paper. So far, we have tried increasing causal or non-causal encoder and found increasing non-causal is more effective. We have also tried deliberation based on cascaded encoders of other sizes (maintaining a total size of 500M) but did not obtain any significant improvement.

\subsection{Comparison}
\label{sec:comparison}

In Table \ref{tab:wer}, we compare the multilingual cascaded encoder baseline (\texttt{B1}) to a deliberation model (\texttt{E3}), and also their large-scale versions: a 1B cascaded encoder model (\texttt{B2}) v.s. a 1B deliberation model with a large text encoder (\texttt{E6}) or 1B deliberation model with a large-scale transformer decoder (\texttt{E8}). Note that the cascaded encoder model (\texttt{B1}) is exactly the first-pass model used for deliberation (\texttt{E3}, \texttt{E6}, and \texttt{E8}).

First, we see that a 300M multilingual deliberation model (\texttt{E3}) improves the average WER of 9 languages from 7.70\% to 7.39\%. The improvement is around 4\% relative, with certain languages such as it-IT up to 6\%. Compared to monolingual deliberation \cite{hu2022improving}, the improvement of multilingual deliberation here seems to be similarly significant. This shows that with sufficient capacity the deliberation decoder can reach the same effectiveness to its monolingual counterparts.

Based on the results of previous sections, we then further increase the deliberation model capacity by increasing either the text encoder (\texttt{E6}) or the transformer decoder (\texttt{E8}). We see in Table \ref{tab:wer} that the deliberation model with a 32L 1080D transformer decoder (\texttt{E8}) performs the best, reducing the average WER by around 9\% relative (from 7.70\% to 6.99 \%) compared to the cascaded encoder baseline (\texttt{B1}). Compared to a 1B multilingual transducer with a 23L 1152D cascaded encoder (\texttt{B2}), the aforementioned 1B deliberation model still performs around 4\% relatively better, demonstrating the effectiveness of increasing the deliberation transformer decoder. We note that our deliberation decoder is based on the transformer layers here and can thus be parallelized during rescoring similar to \cite{li2020parallel}. This significantly reduces the rescoring latency compared to other rescorers based on long short-term memory (LSTM).

Besides using a 23L 1152D cascaded encoder, we have also tried changing the structure of the cascaded encoder in other ways (while maintaining the 1B size)\replaced{ by }{, e.g. }using a deeper or wider encoder, but did not obtain better results. When we increase the cascaded encoder to a 35L 1024D one, the model has trouble converging during training. Lastly, we observe in Table \ref{tab:wer} that increasing the text encoder of the deliberation model (\texttt{E6}) seems to introduce regression. This is because a big text encoder becomes hard to train without pretraining, similar to the observation in~\cite{hu2022improving}. In future, we will research pretraining of multilingual text encoder for deliberation.

\section{Conclusion}
\added{In this work we investigated deliberation for the multilingual cascaded encoder.} We show that deliberation is effective in multilingual setup including 9 languages, with 5\% and 4\% relative average WER improvements for causal and non-causal first-pass models, respectively. Increasing the capacity of the deliberation model is necessary for uniform improvements for all languages. \added{We show that it is more effective to increase the deliberation transformer decoder size than the size of deliberation text encoder or the first-pass cascaded encoder.} Compared to the baseline multilingual cascaded encoder, our 1B deliberation model improves WER by 9\% on average for 9 languages. Compared to a 1B multilingual cascaded encoder, the deliberation model improves the average WER by 4\% relative and performs uniformly better for all languages, with improvements up to 7\%. Our large-scale multilingual deliberation decoder is based on transformer layers and can thus be parallelized across tokens during rescoring. In future, we plan to investigate using large-scale multilingual text-only data to pretrain the deliberation text encoder.

\bibliographystyle{IEEEbib}
\bibliography{refs}

\end{document}